\documentclass{article}
\usepackage{spconf}

\usepackage[T1]{fontenc}
\usepackage{amsmath}
\usepackage{newtxmath}
\usepackage{bm}
\usepackage{graphicx,booktabs,array,tabularx}
\usepackage{xcolor}
\usepackage{microtype}
\usepackage{cite}
\definecolor{ink}{HTML}{20364B}
\definecolor{teal}{HTML}{267C80}
\definecolor{blue}{HTML}{416F9F}
\definecolor{coral}{HTML}{BE6B56}
\usepackage[breaklinks,colorlinks,allcolors=blue]{hyperref}
\hypersetup{pdftitle={DuplexTune: Coupled-State Replay for Memory-Efficient Mixture-of-Experts Adaptation},pdfauthor={Paul Raj; Simon Caldwell; Daniel Whitmore; Mutlu Mete}}
\newcommand{\method}{DuplexTune}
\newcommand{\R}{\mathbb{R}}
\newcommand{\Tin}{T_{\mathrm{in}}}
\newcommand{\Tout}{T_{\mathrm{out}}}
\newcommand{\norm}[1]{\lVert#1\rVert}
\newcommand{\va}{\bm{a}}
\newcommand{\vb}{\bm{b}}
\newcommand{\vu}{\bm{u}}
\newcommand{\vstate}{\bm{v}}
\newcommand{\vr}{\bm{r}}
\newcommand{\Norm}{\operatorname{Norm}}
\newcommand{\Attn}{\operatorname{Attn}}
\newcommand{\MoE}{\operatorname{MoE}}
\newcolumntype{Y}{>{\raggedright\arraybackslash}X}

\title{DuplexTune: Coupled-State Replay for\\Memory-Efficient Mixture-of-Experts Adaptation}

\name{Paul Raj\textsuperscript{1}\qquad
Simon Caldwell\textsuperscript{2}\qquad
Daniel Whitmore\textsuperscript{2}\qquad
Mutlu Mete\textsuperscript{1}}
\address{\textsuperscript{1}Texas A\&M University-Commerce\\
\textsuperscript{2}Michigan State University}

\begin{document}
\maketitle

\begin{abstract}
Sparse expert execution does not remove the stored states needed to adapt a mixture-of-experts language model. We present \method, a framework for replaying coupled hidden states through pretrained attention and expert modules. Trainable width bridges connect two residual streams to the backbone. Bridge warm-up precedes backbone adaptation, with router weights kept frozen. We analyze the implicit state-recovery equation and identify the convergence and replay conditions needed for reconstruction. On Qwen1.5-MoE-A2.7B, the single-H800 evaluation reports 38.8\,GB peak memory, versus 64.8\,GB for supervised fine-tuning with checkpointing, and throughput of 25.1 versus 19.9 samples/s. \method{} achieves 67.2\% MMLU, 75.6\% GSM8K, and 7.69 MT-Bench, with a weaker multilingual score than the checkpointed baseline. The results characterize a configuration-dependent tradeoff among state storage, throughput, and adaptation quality.
\end{abstract}

\section{Introduction}
MoE activation is sparse, but training still maintains weights, gradients, optimizer states, and activations~\cite{DBLP:journals/corr/ShazeerMMDLHD17,fedus2022switchtransformersscalingtrillion}. Qwen1.5-MoE-A2.7B contains 14.3 billion parameters but activates approximately 2.7 billion~\cite{qwen2024moe}. Active computation and stored-state costs therefore need separate accounting when adapting Transformer language models~\cite{DBLP:journals/corr/VaswaniSPUJGKP17,openai2024gpt4technicalreport,touvron2023llamaopenefficientfoundation}.

LoRA restricts trainable updates, while QLoRA also quantizes the frozen backbone~\cite{DBLP:journals/corr/abs-2106-09685,dettmers2023qloraefficientfinetuningquantized}. ZeRO and FSDP shard model states~\cite{rajbhandari2020zero,zhao2023pytorchfsdpexperiencesscaling}, whereas checkpointing recomputes intermediate features on single or multiple devices~\cite{DBLP:journals/corr/ChenXZG16}.

\method{} frames this target as \emph{coupled-state replay}. Two half-width residual streams exchange information through pretrained attention and expert modules. Learnable width bridges connect the streams to those modules. Bridge warm-up establishes the interface before backbone adaptation updates attention, expert weights, and bridges while leaving router weights fixed. This expands learning beyond adapters without describing the frozen-router configuration as an update to every parameter.

Because attention queries depend on the stream being recovered, replay requires implicit reconstruction. We state a sufficient convergence condition, a residual error bound, and replay requirements. The evaluation compares complete configurations: 38.8 versus 64.8\,GB for checkpointed SFT, with higher scores on three tasks but weaker multilingual performance.

\begin{figure*}[t]
\centering
\includegraphics[width=\textwidth]{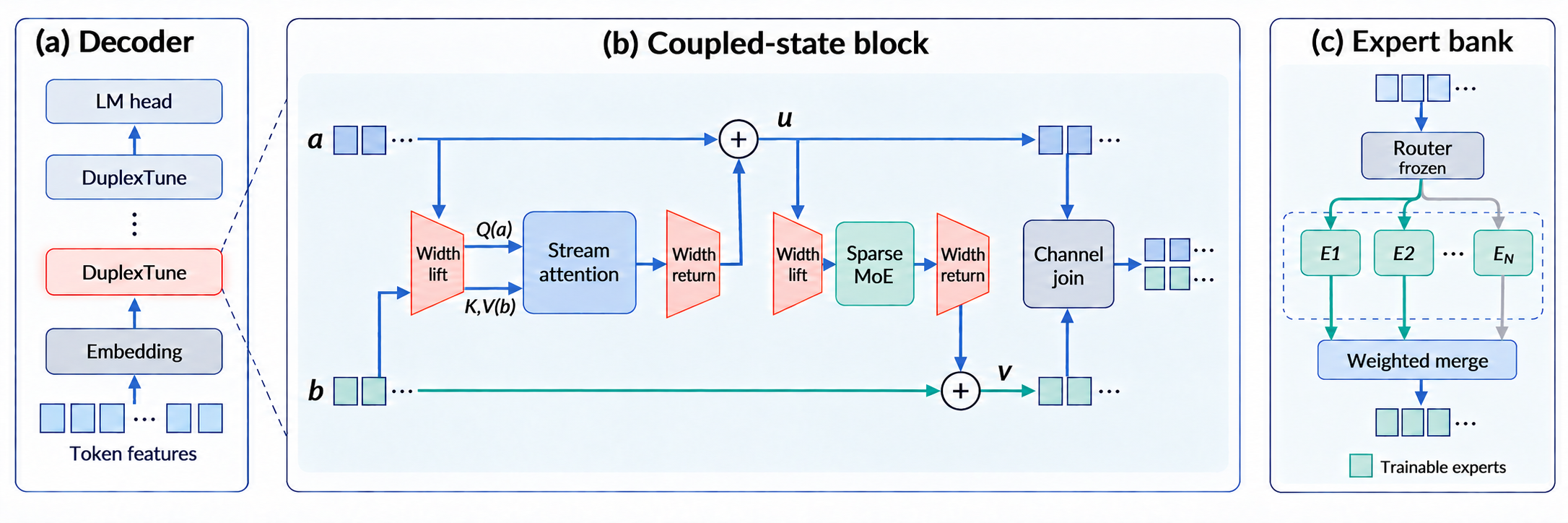}
\caption{\textbf{DuplexTune architecture.} (a) Adapted decoder stack. (b) Width-lift and width-return bridges connect paired states to pretrained attention and sparse experts. Attention updates $\va$ to $\vu$; expert processing then updates $\vb$ to $\vstate$ before the channel join. (c) Schematic expert bank with frozen router weights during backbone adaptation.}
\label{fig:architecture}
\end{figure*}

\section{Related Work}
\noindent\textbf{Routing and resource allocation.}
MoE surveys distinguish expert architecture, routing, and system execution~\cite{cai2025moesurvey}. At the agent level, KABB and GAM-Agent coordinate expertise~\cite{z1,z2}. Training-system reviews separately examine memory placement, parallelism, and communication~\cite{duan2024trainingsurvey}. Agora trades uncertainty, while Markets of Minds evaluates assistance through marginal benefit and opportunity cost~\cite{zhang2026doubtsyourselftradingvisual,zhang2026marketsofminds}. Agent allocation and token-level state recovery operate at different granularities.

\noindent\textbf{Adaptation objectives and optimization.}
PEFT surveys organize adaptation by the trainable parameterization~\cite{han2024peftsurvey}. CF-VLM and VLM learning dynamics address adaptation objectives~\cite{CF,z10}. LoRA, DoRA, and $(\mathrm{IA})^3$ restrict trainable updates~\cite{DBLP:journals/corr/abs-2106-09685,liu2024doraweightdecomposedlowrankadaptation,liu2022fewshotparameterefficientfinetuningbetter}. Self-rewarded multimodal reasoning studies adaptation signals, while MAT-Agent organizes training optimization~\cite{zhang2025srmcr,z4}. LOMO reduces optimization memory, GaLore projects gradients, and Adam determines optimizer-state costs~\cite{lv2024parameterfinetuninglargelanguage,zhao2024galorememoryefficientllmtraining,kingma2017adammethodstochasticoptimization}. OSC aligns collaborator knowledge, and failure-driven refinement organizes workflows~\cite{z9,z8}. These objectives and coordination mechanisms differ from activation replay.

\noindent\textbf{Reconstruction and memory budgets.}
RevNet uses triangular coupling, and Reformer brings reversible layers to Transformers~\cite{gomez2017reversible,DBLP:journals/corr/abs-2001-04451}. Invertible residual networks recover states under a contraction condition~\cite{behrmann2019invertible}, relevant to our query-dependent attention. Routing also appears in diffusion~\cite{z11}, targeting a different computation--quality tradeoff. FlashAttention reduces memory traffic through IO-aware exact attention, whereas ZeRO-Offload moves state and computation to the CPU~\cite{dao2022flashattention,ren2021zerooffload}. HTC-VLM compresses visual tokens; video encoders built on frozen image representations allocate tokens before temporal refinement~\cite{zhang2025htcvlm,zhang2026videoencoders}. These mechanisms change different costs from retaining hidden states across layers.

\noindent\textbf{Evaluation scope.}
MMLU, GSM8K, and MT-Bench assess knowledge, mathematical reasoning, and conversational responses~\cite{DBLP:journals/corr/abs-2009-03300,cobbe2021training,zheng2023judging}. Tri-MARF and MM-CoT instead examine annotation and visual reasoning~\cite{Z5,z6}. The multimodal and multi-agent studies above provide context for adaptation and allocation; they are not memory baselines in our language-model evaluation.

\section{Coupled-State Replay with Width Bridges}
\label{sec:method}

\subsection{Paired States and Width Bridges}
Let $\bm H=[\va,\vb]\in\R^{B\times S\times d}$, where each stream has width $d/2$, $B$ is batch size, and $S$ is sequence length. The transition pair
\begin{equation}
\begin{aligned}
\Tin&\colon\R^{d/2}\longrightarrow\R^d,\\
\Tout&\colon\R^d\longrightarrow\R^{d/2}
\end{aligned}
\label{eq:transitions}
\end{equation}
connects these streams to the pretrained sublayers. We call $\Tin$ the width-lift bridge and $\Tout$ the width-return bridge. With stream normalization $\Norm$, the projected states are
\begin{equation}
\begin{aligned}
\bar{\va}&=\Tin\bigl(\Norm(\va)\bigr),&
\bar{\vb}&=\Tin\bigl(\Norm(\vb)\bigr).
\end{aligned}
\label{eq:project}
\end{equation}
The wrapped attention and expert functions are
\begin{equation}
\begin{aligned}
\mathcal A(\va,\vb)
 &=\Tout\bigl(\Attn_{\mathrm{pt}}(\bar{\va},\bar{\vb},\bar{\vb})\bigr),\\
\mathcal M(\va)
 &=\Tout\bigl(\MoE_{\mathrm{pt}}(\Tin(\Norm(\va)))\bigr).
\end{aligned}
\label{eq:modules}
\end{equation}
Attention uses queries from $\va$ and keys and values from $\vb$, with a causal mask. The coupled update is
\begin{equation}
\begin{aligned}
\vu&=\va+\mathcal A(\va,\vb),\\
\vstate&=\vb+\mathcal M(\vu).
\end{aligned}
\label{eq:forward}
\end{equation}
The next layer receives $[\vu,\vstate]$. Attention and expert operations retain their pretrained matrix dimensions, while residual streams remain half-width.

For a shared transition pair, the additional weight count is $d^2$ per layer, excluding biases; independent pairs for attention and MoE would require $2d^2$. Preserving pretrained matrix shapes does not make the wrapper functionally identical to the original decoder. In particular, the rank of the linear stream interface is at most $d/2$. Alignment must therefore learn a useful interface rather than assume lossless recovery of the original representation.

\subsection{Conditional State Replay}
Given $\vu$ and $\vstate$, the second input stream follows directly:
\begin{equation}
\hat{\vb}=\vstate-\mathcal M(\vu).
\label{eq:recover_b}
\end{equation}
The first stream satisfies $\va=\vu-\mathcal A(\va,\vb)$. Unlike triangular RevNet coupling, $\va$ also appears inside the attention query. Its reconstruction uses
\begin{equation}
\begin{aligned}
\va^{(0)}&=\Norm(\vu),\\
\va^{(j+1)}&=\vu-\mathcal A(\va^{(j)},\hat{\vb}),\qquad j\geq0.
\end{aligned}
\label{eq:iterate}
\end{equation}
This equation is a reconstruction procedure, not an unconditional analytic inverse.

For fixed $\vb$, suppose $\va\mapsto\vu-\mathcal A(\va,\vb)$ maps a closed domain into itself and is Lipschitz with constant $q<1$. This gives a unique fixed point, as in the contraction argument for invertible residual networks~\cite{behrmann2019invertible}. For a candidate $\hat{\va}$ and exact $\vb$, the residual and error bound are
\begin{equation}
\begin{aligned}
\vr&=\hat{\va}+\mathcal A(\hat{\va},\vb)-\vu,\\
\norm{\hat{\va}-\va}&\leq\frac{\norm{\vr}}{1-q}.
\end{aligned}
\label{eq:residualbound}
\end{equation}
The bound follows by subtracting the true fixed-point equation and applying the triangle inequality. A small residual is informative only when the contraction factor is controlled. Additive coupling alone does not establish this condition, and finite-precision replay need not be exact.

\noindent\textbf{Local invertibility.}
Within a differentiable region of fixed expert dispatch, let $\bm J_a=\partial\mathcal A/\partial\va$. The forward map factors into $(\va,\vb)\mapsto(\vu,\vb)$ and $(\vu,\vb)\mapsto(\vu,\vstate)$. The second factor has unit Jacobian determinant, giving
\begin{equation}
\det\!\left(\frac{\partial(\vu,\vstate)}{\partial(\va,\vb)}\right)
 =\det\bigl(\mathbf I+\bm J_a\bigr).
\label{eq:jacobian}
\end{equation}
Local invertibility therefore depends on query-dependent attention, although recovering $\vb$ is a direct subtraction. Contraction keeps $\mathbf I+\bm J_a$ nonsingular, but must be verified for the learned functions and numerical precision.

Backward replay requires the same parameters, attention masks, random states, and expert dispatch decisions as the forward computation. Frozen router weights do not freeze token assignments when inputs change. Gradients must follow the replayed forward block rather than an arbitrary unrolling of inverse iterations.

Approximate state recovery perturbs the gradient evaluation points. Reconstruction residuals and gradient agreement with a stored-activation reference are therefore separate numerical checks; downstream task scores alone validate neither.

\subsection{Bridge Warm-Up and Backbone Adaptation}
Let $\theta$ denote pretrained parameters, $\theta_r$ the router subset, and $\phi$ the transitions. Autoregressive supervised adaptation minimizes
\begin{equation}
\mathcal L(\theta,\phi)
=-\sum_{(c,y)}\sum_j\log p_{\theta,\phi}(y_j\mid c,y_{<j}).
\label{eq:loss}
\end{equation}
During \emph{bridge warm-up}, Phase~1 freezes $\theta$ and updates only $\phi$. Phase~2, \emph{backbone adaptation}, optimizes $\phi$ and $\theta\setminus\theta_r$. Attention and expert weights become trainable while router weights stay fixed. This extends learning beyond the bridges without updating every parameter. The ablation evaluates the schedule, without assigning its gains to an unmeasured convergence mechanism.

\subsection{Resource Accounting}
Peak GPU memory is the maximum concurrent storage of weights, gradients, optimizer states, activations, and workspace. Reconstruction targets activations; stored expert weights and optimizer states remain. A replayable segment can reduce stored layer-boundary states from $O(LBSd)$ to $O(BSd)$ for depth $L$, but attention workspace, temporary expert activations, and replay metadata remain. Fixed-point attention recovery adds evaluations whose count depends on numerical convergence.

Sparse computation and state residency remain distinct: selecting fewer experts need not reduce stored weights. Markets of Minds addresses shared-capacity allocation~\cite{zhang2026marketsofminds}; our frozen-router wrapper instead targets activation storage. Precision, optimizer representation, and state placement still determine total memory. The evaluation compares complete configurations, without separately profiling activation savings or implementing market clearing.

\section{Experiments}
\subsection{Evaluation Setting}
The evaluation adapts \texttt{Qwen/Qwen1.5-MoE-A2.7B} on \texttt{databricks-dolly-15k}, using PyTorch and Transformers on one 80\,GB NVIDIA H800. The checkpoint uses the \texttt{qwen2\_moe} implementation~\cite{qwen2024moe}.

The comparison includes LoRA, DoRA, and $(\mathrm{IA})^3$ as parameter-efficient methods, and supervised fine-tuning (SFT) with checkpointing, LOMO, and GaLore as methods with broader weight updates. Peak GPU memory is reported in GB and throughput in samples/s. The methods use the same number of training steps, with batch sizes adjusted to the memory capacity. This setting measures achievable throughput for each configuration; it does not control processed tokens or isolate kernel efficiency.

Task metrics are MMLU~\cite{DBLP:journals/corr/abs-2009-03300}, GSM8K~\cite{cobbe2021training}, and MT-Bench~\cite{zheng2023judging}, together with the multilingual aggregate in the comparison. The multilingual dataset composition is unspecified, so that column is interpreted within this evaluation rather than as a standardized external benchmark. The available results contain no multi-seed uncertainty estimates.

\begin{table}[t]
\centering
\caption{Single-H800 efficiency. Batch sizes are capacity-tuned; throughput is not a matched-token comparison.}
\label{tab:memory_speed}
\small
\begin{tabular*}{\columnwidth}{@{\extracolsep{\fill}}lrr@{}}
\toprule
Method & Memory (GB)$\downarrow$ & Samples/s$\uparrow$\\
\midrule
\multicolumn{3}{@{}l}{\emph{Parameter-efficient adaptation}}\\
LoRA & 18.4 & \textbf{76.2}\\
DoRA & 19.8 & 72.5\\
$(\mathrm{IA})^3$ & \textbf{17.6} & 74.8\\
\midrule
\multicolumn{3}{@{}l}{\emph{Broader weight updates}}\\
SFT + checkpointing & 64.8 & 19.9\\
LOMO & 41.9 & 17.6\\
GaLore & 44.8 & 35.5\\
\textbf{\method} & 38.8 & 25.1\\
\bottomrule
\end{tabular*}
\end{table}

\begin{table}[t]
\centering
\caption{Downstream scores. MMLU, GSM8K, and Multi. are percentages; MT-B. is MT-Bench. Bold marks the best score per column, including the base model.}
\label{tab:performance}
\small
\begin{tabular*}{\columnwidth}{@{\extracolsep{\fill}}lrrrr@{}}
\toprule
Method & MMLU & GSM8K & Multi. & MT-B.\\
\midrule
Base model & 62.4 & 61.2 & \textbf{40.4} & 6.25\\
\midrule
LoRA & 65.5 & 71.8 & 38.7 & 7.21\\
DoRA & 65.9 & 71.1 & 39.1 & 7.29\\
$(\mathrm{IA})^3$ & 65.2 & 70.5 & 38.4 & 7.18\\
\midrule
SFT + checkpointing & 66.3 & 75.1 & 39.8 & 7.56\\
LOMO & 66.5 & 74.9 & 39.5 & 7.52\\
GaLore & 66.6 & 74.5 & 39.4 & 7.49\\
\textbf{\method} & \textbf{67.2} & \textbf{75.6} & 39.0 & \textbf{7.69}\\
\bottomrule
\end{tabular*}
\end{table}

\subsection{Memory--Throughput Tradeoff}
Table~\ref{tab:memory_speed} reports 38.8\,GB peak memory for \method{}, versus 64.8\,GB for checkpointed SFT. The reduction is 26.0\,GB, or 40.1\%. Throughput increases from 19.9 to 25.1 samples/s, a 26.1\% gain under the reported batching policy. This difference cannot establish that reconstruction itself is faster than checkpointing: batch sizes and other configuration choices contribute to the measured speed.

\method{} uses 3.1\,GB less memory than LOMO and 6.0\,GB less than GaLore, but is 29.3\% slower than GaLore's 35.5 samples/s. PEFT remains more memory efficient: $(\mathrm{IA})^3$ uses 17.6\,GB, and LoRA reaches 76.2 samples/s. These configurations offer different tradeoffs between update scope, memory, and throughput.

\subsection{Adaptation Quality}
Table~\ref{tab:performance} shows the highest adapted-model scores for \method{} on MMLU, GSM8K, and MT-Bench. Relative to checkpointed SFT, the gains are 0.9 and 0.5 percentage points on MMLU and GSM8K, and 0.13 MT-Bench points. Against DoRA, the corresponding differences are 1.3, 4.5, and 0.40. These are descriptive differences from the reported runs; the table does not establish statistical significance.

Multilingual performance shows a different pattern: \method{} scores 39.0, below checkpointed SFT (39.8), DoRA (39.1), and the base model (40.4). Every adapted model underperforms the base in this column. Reasoning and conversational gains therefore coexist with a lower multilingual score.

\begin{figure}[t]
\centering
\includegraphics[width=\columnwidth]{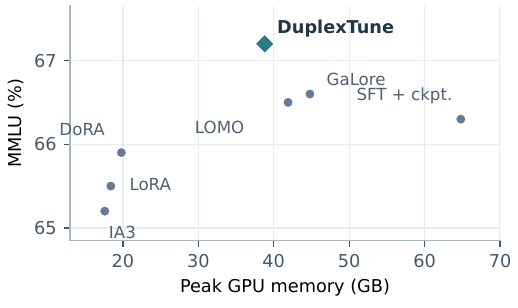}
\caption{\textbf{Memory and MMLU.} Points combine the values in Tables~\ref{tab:memory_speed} and~\ref{tab:performance}; they summarize complete configurations with capacity-tuned batches. Throughput and other task scores remain separate criteria.}
\label{fig:tradeoff}
\end{figure}

\begin{table}[t]
\centering
\caption{Training-phase ablation on MMLU. Removing either alignment or backbone fine-tuning reduces the score.}
\label{tab:ablation}
\small
\begin{tabular*}{\columnwidth}{@{\extracolsep{\fill}}lr@{}}
\toprule
Configuration & MMLU (\%)$\uparrow$\\
\midrule
\textbf{\method{} (complete)} & \textbf{67.2}\\
No bridge warm-up (joint updates) & 57.5\\
No backbone adaptation (bridges only) & 54.9\\
\bottomrule
\end{tabular*}
\end{table}

\subsection{Training-Phase Ablation}
Table~\ref{tab:ablation} compares the complete schedule with removing either phase. Omitting alignment yields 57.5\% MMLU, 9.7 percentage points below the complete method. Restricting learning to transitions yields 54.9\%, a 12.3-point decrease. Both variants also fall below the 62.4\% base-model score. These observations support using alignment and subsequent backbone updates together in this configuration. They do not, without optimization traces, prove that the joint-training variant diverges or identify the mechanism behind the decline.

\subsection{Scope and Reproducibility}
The evaluation covers one backbone, one instruction dataset, and one GPU class. Reconstruction, transitions, router freezing, and optimizer-state storage are not separately ablated. Memory comparisons depend on precision and state placement. Reproduction additionally requires learning rates, sequence lengths, per-method batches, inverse-solver tolerances, and multilingual task composition. The H800 results do not establish single-consumer-GPU feasibility.

Token compression and state replay also act on different budgets. HTC-VLM and video encoders built on image representations reduce visual-token exposure~\cite{zhang2025htcvlm,zhang2026videoencoders}, whereas this evaluation changes hidden-state storage. Combining them would require measuring retained task evidence and reconstruction error alongside memory. Markets of Minds~\cite{zhang2026marketsofminds} allocates expert assistance by expected benefit and opportunity cost; that decision does not determine an expert's training-state footprint.

\section{Conclusion}
\method{} combines coupled-state replay, width bridges, and staged backbone adaptation. State recovery requires convergence; weight and optimizer costs remain. The reported H800 configuration reduces memory relative to checkpointed SFT and improves three task scores, with multilingual and throughput tradeoffs. Further evaluation should measure matched-token quality, reconstruction error, and state residency\label{maintextend}.%

\end{document}